\documentclass[12pt]{article}
\usepackage{arxiv}
\usepackage{algorithm,algorithmic}
\usepackage{amsmath,amssymb,amsfonts}
\usepackage{todonotes}
\usepackage{float}
\usepackage{multicol,multirow}
\usepackage{secdot}
\usepackage{makecell}
\usepackage{amsmath,array,ragged2e}
\newcolumntype{C}{>{\Centering\arraybackslash}m{0.14\linewidth}}
\usepackage{booktabs}
\usepackage{adjustbox}
\usepackage{subcaption}
\usepackage{graphicx}
\usepackage{hyperref}

\usepackage[numbers]{natbib}  

\usepackage{pifont}  
\newcommand{\xmark}{\textcolor{red}{\ding{55}}}
\newcommand{\cmark}{\textcolor{green!60!black}{\ding{51}}}

\usepackage{threeparttable}  

\usepackage{bm}  

\usepackage{balance} 

\usepackage{amsthm}

\renewcommand{\shorttitle}{UQ of Click and Conversion Estimates for Autobidding}

\date{}
\begin{document}

\title{Uncertainty Quantification of Click and Conversion Estimates for the Autobidding}

\author{
    Ivan Zhigalskii\\
    Avito \& HSE\\
    Moscow, Russia\\
    \texttt{iazhigalskiy@edu.hse.ru}
\And 
    Andrey Pudovikov\\
    AI Center \& IAI MSU \\
    Lomonosov Moscow State University \\ 
    Moscow, Russia\\
    \texttt{pudovikovad@my.msu.ru}\\
\And
    Aleksandr Katrutsa\\
    Avito \& IAI MSU\\
    Moscow, Russia\\
    \texttt{amkatrutsa@gmail.com}
\And 
    Egor Samosvat\\
    Avito\\
    Moscow, Russia\\
    \texttt{easamosvat@avito.ru}
}

\maketitle
\begin{abstract}
Modern e-commerce platforms employ various auction mechanisms to allocate paid slots for a given item.
To scale this approach to the millions of auctions, the platforms suggest promotion tools based on the autobidding algorithms.
These algorithms typically depend on the Click-Through-Rate (CTR) and Conversion-Rate (CVR) estimates provided by a pre-trained machine learning model.
However, the predictions of such models are uncertain and can significantly affect the performance of the autobidding algorithm.
To address this issue, we propose the \texttt{DenoiseBid} method, which corrects the generated CTRs and CVRs to make the resulting bids more efficient in auctions.
The underlying idea of our method is to employ a Bayesian approach and replace noisy CTR or CVR estimates with those from recovered distributions.
To demonstrate the performance of the proposed approach, we perform extensive experiments on the synthetic, iPinYou, and BAT datasets.
To evaluate the robustness of our approach to the noise scale, we use synthetic noise and noise estimated from the predictions of the pre-trained machine learning model.
\end{abstract}

\section{Introduction}
\label{sec:intro}

Modern e-commerce platforms allocate advertising slots through automated auction mechanisms that process millions of auctions daily. 
To operate at this scale, platforms deploy autobidding systems that compute optimal bids for advertisers.
These systems solve constrained optimization problems parameterized by Click-Through Rates~(CTR) and Conversion Rates~(CVR), which are estimated from data.
Under the widely studied Second-Price Auction~(SPA) rule~\cite{yang2019bid, aggarwal2024auto, he2021unified, zhang2017managing, khirianova2025bat, pudovikov2025autobidding, pudovikov2025robust}, the optimization problem for autobidding systems can be stated as a linear programming~(LP) problem.
From such a problem follows that the optimal bid is a linear function of CTR and CVR~\cite{yang2019bid, aggarwal2024auto}.
Consequently, estimation errors in these quantities propagate directly into the bid, potentially leading to suboptimal budget allocation and violations of the advertiser's cost-per-click~(CPC) constraints.
We focus on SPA because the resulting linear dependence of bid on external quantities, e.g., CTR and CVR, allows Bayesian treatment of uncertainties in their estimates.
The extension to first-price auctions is an important direction that requires alternative approaches and is left for future work.

In practice, CTR and CVR are predicted by machine learning models whose outputs are inherently uncertain.
This motivates the development of a bidding framework that explicitly accounts for predictive noise.
We propose \texttt{DenoiseBid}, a Bayesian autobidding method that replaces noisy estimates of CTR and CVR with their posterior expectations, conditioned on the model's predictions and a prior distribution recovered from historical data.

The main contributions of this study are as follows:
\begin{enumerate}
    \item We state the autobidding problem under the noisy CTR and CVR values and derive a closed-form bidding rule based on their posterior expectations.
    
    \item We develop the \texttt{DenoiseBid} method, which recovers the prior distribution of CTR and CVR from observations and computes denoised bids in closed form.
    
    \item We perform extensive empirical validation of \texttt{DenoiseBid} on four datasets under both synthetic and empirical noise and demonstrate consistent gains.
\end{enumerate}


\section{Related works.}


\paragraph{Autobidding formulations.}
The autobidding problem can be viewed as a specialization of Real-Time Bidding~(RTB)~\cite{ou2023survey} in which the platform optimizes bids under global performance goals.
Existing approaches employ optimization~\cite{aggarwal2024auto, zhang2017managing}, reinforcement learning~\cite{cai2017real, he2021unified}, or control techniques~\cite{yang2019bid, stram2024mystique}.
To the best of our knowedge, \cite{yang2019bid} were the first to cast the advertiser's conversion-maximization problem with joint budget and CPC constraints as an LP; the formulation and its dual-based bidding rule have since become a standard baseline~\cite{aggarwal2024auto, pudovikov2025robust}.

\paragraph{Uncertainty quantification for CTR/CVR 
prediction.} 
The LP-based bids depend on estimated CTR and CVR, and the autobidding problem can itself be viewed through a Bayesian lens~\cite{aggarwal2024auto}.
From the modeling side, CTR/CVR prediction spans tabular feature-interaction models and deep behavior-sequence architectures~\cite{zhu2022bars}.
For gradient-boosted decision trees~(GBDT), virtual ensembles decompose predictive variability into data and knowledge components~\cite{malinin2020uncertainty}.
The similar ideas for deep models include deep ensembles and variational Bayesian methods~\cite{abdar2021review}.
Hence, calibrated uncertainty estimates are available for the dominant model families used in practice.

\paragraph{Propagating uncertainty into bidding.} 
The critical remaining question is how to incorporate these calibrated uncertainties into the autobidding layer.
\cite{shih2023robust} cluster CTR values and combine them with reinforcement learning to construct a bidding policy; however, their clustering-based discretization can lose information when the CTR distribution is unimodal or highly concentrated.
\cite{zhang2017managing} introduce a risk-management framework that models CTR uncertainty via Bayesian logistic regression; while principled, this restricts the prediction model to a specific parametric family, precluding the use of modern GBDT or deep-learning predictors~\cite{zhu2022bars}.
The recent \texttt{RobustBid}~\cite{pudovikov2025robust} applies robust optimization with uncertainty sets constructed from quantile estimates of CTR and CVR distributions, achieving strong constraint satisfaction; however, it does not incorporate per-prediction uncertainty estimates from the ML model.



\texttt{DenoiseBid} jointly leverages the empirical distribution of CTR and CVR values and per-prediction uncertainty estimates to construct denoised bids (see Table~\ref{table:comparison}).
Unlike prior methods, it is model-free, retains a closed-form bidding rule, and exploits Bayesian posterior expectations.

\begin{table}[!h]
  \caption{Comparison \texttt{DenoiseBid} with other methods.
  }
  \label{table:comparison}
  \centering
  \begin{tabular}{cccc}
    \toprule
    \textbf{Reference} 
      & \textbf{\makecell{Multiple \\ uncertainty sources}} 
      & \textbf{\makecell{Usage of CTR \\ distribution}} 
      & \textbf{\makecell{UQ of prediction \\ model}} \\
    \midrule
    \citep{shih2023robust} 
      & \xmark 
      & \cmark 
      & \xmark \\
    \addlinespace
    \citep{zhang2017managing} 
      & \xmark 
      & \cmark 
      & \xmark \\
    \addlinespace
    \citep{pudovikov2025robust} 
      & \cmark 
      & \cmark 
      & \xmark \\
    \addlinespace
    \textbf{This paper} 
      & \cmark 
      & \cmark 
      & \cmark \\
    \bottomrule
  \end{tabular}
\end{table}


\section{Problem statement}
\label{sec:problem}

This section formalizes the deterministic autobidding optimization problem under perfect knowledge of click and conversion probabilities. 
The resulting bidding formula serves as the baseline that Section~\ref{sec:denoise-bid} extends to the Bayesian setting.

Consider a single advertiser participating in a sequence of~$T$ single-slot second-price auctions (SPAs). 
For each auction $t \in [T] = \{1, \dots, T\}$, we denote by $CTR_t$ the click-through rate, by $CVR_t$ the conversion rate, by $wp_t > 0$ the realized winning price, and by $bid_t \geq 0$ the advertiser's bid. 
We adopt an offline setting in which the winning prices $\{wp_t\}_{t=1}^T$ and the ground-truth probabilities $\{CTR_t, CVR_t\}_{t=1}^T$ are known. 
Under SPA rules, the 
advertiser wins auction~$t$ whenever $bid_t \geq wp_t$ and pays $wp_t$.



The advertiser seeks to maximize the total expected number of conversions subject to a budget constraint~$B$ and a target cost-per-click (CPC) limit~$C$, both given in advance. 
Following~\cite{yang2019bid, aggarwal2024auto}, we relax the binary allocation variables $x_t \in \{0,1\}$ to $x_t \in [0, 1]$, yielding the following LP problem:
\begin{equation}
    \begin{split}
        & \max_{0 \leq x_t \leq 1} \quad 
        \sum_{t=1}^T x_t \cdot CTR_t \cdot CVR_t \\
        \text{s.t. } & 
        \sum_{t=1}^T x_t \cdot wp_t \leq B, \\
        &\frac{\sum_{t=1}^T x_t \cdot wp_t}
              {\sum_{t=1}^T x_t \cdot CTR_t} \leq C.
    \end{split}
    \label{eq::nonrobust_problem}
\end{equation}
Note that it is possible to extend the CPC constraint to a more general case with Return On Investment constraints~\cite{he2021unified}; however, the CPC formulation is more widespread in the literature~\cite{aggarwal2024auto}.

The optimal bid is derived by forming the Lagrangian dual problem
of~\eqref{eq::nonrobust_problem} with dual variables $p \geq 0$ and $q \geq 0$ corresponding to budget and CPC constraints, and applying complementary slackness conditions. 
This yields the closed-form expression
\begin{equation}
    bid_t = \frac{1}{p^*+q^*}\, CVR_t \cdot CTR_t 
          + \frac{q^*}{p^*+q^*}\, C \cdot CTR_t,
    \label{eq::non_robust_bid}
\end{equation}
where $p^*$ and $q^*$ are the optimal dual variables. 
The detailed derivation follows the same scheme as in~\cite{yang2019bid, 
aggarwal2024auto} and is extended to stochastic modification of problem~(\ref{eq::nonrobust_problem}) in Section~\ref{sec:denoise-bid}.

Equation~\eqref{eq::non_robust_bid} shows that $bid_t$ is a linear function of $CTR_t$ and $CVR_t$, meaning that estimation errors in these quantities propagate directly into the bid and can affect the performance of the bidding strategy. 
Section~\ref{sec:denoise-bid} extends the optimization problem~\eqref{eq::nonrobust_problem} to handle such 
uncertainty from the Bayesian perspective.

\section{\texttt{DenoiseBid} method}
\label{sec:denoise-bid}
We assume the bidding agent observes noisy estimates of the ground-truth click and conversion probabilities. 
However, the agent’s objective is to maximize the actual realized conversions.
Therefore, the core idea of \texttt{DenoiseBid} is to shift from deterministic to a stochastic optimization, maximizing the conditional expectation of the true conversion probability given the noisy observations.
Formalization of this idea leads to the following optimization problem:
\begin{equation}
\label{eq::stochastic_optimization_problem}
    \begin{split}
        \max _{0 \leq x_{t} \leq 1} & \space \mathbb{E} \left[ \sum _{t=1}^T x_{t} \cdot CTR_{t} \cdot CVR_{t} \middle| \mathcal{O} \right],  \\
        \text{s.t. } & \space \quad \sum _{t=1}^T x_{t} \cdot w p_{t} \leq B, \\
        & \space \mathbb{E} \left[ \frac{\sum_{t=1}^T x_{t} \cdot w p_{t}}{\sum_{t=1}^T x_{t} \cdot CTR_{t}} \middle| \mathcal{O} \right] \leq C,
    \end{split}
\end{equation}
where the expectation is taken over the latent random variables $\left\{CTR_t, CVR_t\right\}_{t=1}^T$, conditioned by observations $\mathcal{O} = \left\{ \left( \widehat{CTR}_t, \widehat{CVR}_t \right)\right\}_{t=1}^{T}$.

To reduce the stochastic optimization problem to LP, we employ a first-order Taylor approximation for the expectation of a ratio in the average cost-per-click constraint:
\begin{equation}
    \frac{\sum_{t=1}^T x_{t} \cdot w p_{t}}{\mathbb{E} \left[\sum_{t=1}^T x_{t} \cdot CTR_{t} \middle| \mathcal{O} \right]} \leq C.
\end{equation}  
By rearranging the terms, we arrive at the canonical primal formulation:
\begin{equation}
\label{eq::denoisebid_primal_problem}
    \begin{split}
        &\max_{0 \leq x_{t} \leq 1} \quad \mathbb{E} \left[ \sum _{t=1}^T x_{t} \cdot CTR_{t} \cdot CVR_{t} \middle| \mathcal{O} \right],  \\
        \text {s.t.} \quad & \sum _{t=1}^T x_{t} \cdot w p_{t} \leq B, \\
        & \sum_{t=1}^T x_{t} \cdot \left(wp_{t} - C \cdot\mathbb{E} \left[ CTR_{t} \middle| \mathcal{O} \right] \right) \leq 0.
    \end{split}
\end{equation}
This representation demonstrates that the agent optimizes the posterior mean of conversions while accounting for the target CPC based on the posterior mean of the CTR.

Similarly to Section~\ref{sec:problem}, we use  the Lagrangian dual of problem~\eqref{eq::denoisebid_primal_problem} and complementary slackness conditions to obtain the optimal bidding formula:
\begin{equation}
\label{eq::denoisebid_bid_formula}
\begin{split}
    bid_t = \frac{1}{p^* + q^*} \mathbb{E}[CTR_t \, CVR_t \mid \mathcal{O} ] + \frac{q^*}{p^* + q^*} C \cdot \mathbb{E}[CTR_t \mid \mathcal{O}],
\end{split}
\end{equation}
where $p^*$ and $q^*$ are the optimal dual variables associated with the budget and CPC constraints, respectively.
The full derivation of~(\ref{eq::denoisebid_bid_formula}) is provided in Appendix~\ref{appx::stoc_opt_prob}).

Consequently, the problem of optimal bidding reduces to the computation of conditional expectations. 
Applying Bayes' formula, we write out the expectation of a latent variable~$X$ (representing $CTR$ or $CTR \cdot CVR$) in terms of the likelihood and the prior distribution:
\begin{equation}
\begin{split}
    \mathbb{E}[X \mid \mathcal{O}] = \int_X X \cdot p(X \mid \mathcal{O}) \mathrm{d}X=\frac{\int_X X \cdot p(\mathcal{O} \mid X) p(X) \mathrm{d}X}{\int_X p(\mathcal{O} \mid X) p(X) \mathrm{d}X}.
\end{split}
\end{equation}
Thus, calculating the posterior expectation requires the following components: a noise model (likelihood) $p(\mathcal{O} \mid X)$ and a prior distribution $p(X)$, which captures the distribution of the true values. 
In the subsequent sections, we analyze the calculation of these expectations under different specifications, addressing CTR-only and joint CTR-CVR uncertainty scenarios. 
As a result, we derive expressions for the experimental evaluation of the \texttt{DenoiseBid} method.

\subsection{Bidding under CTR-only Uncertainty}
\label{sec:ctr-uncertainty}
To illustrate the mechanism of the \texttt{DenoiseBid} framework, we first consider a simple case where noise is primarily concentrated in CTR predictions, i.e., $\widehat{CVR}_t = CVR_t$. 
Furthermore, we assume that CTR and CVR predictions are independent of each other and across different auctions. 
Under these assumptions, the joint posterior expectation of value in~\eqref{eq::denoisebid_bid_formula} decomposes as:
\begin{equation}
    \label{eq::value_exp_single_ctr_unc}
    \begin{split}
        \mathbb{E}\biggl[CTR_t \cdot CVR_t \bigg| \Big\{\Big(\widehat{CTR}_t, \widehat{CVR}_t\Big) \Big\}_{t=1}^T \biggr] = CVR_t \cdot \mathbb{E} \left[ CTR_t \middle| \widehat{CTR}_t \right].
    \end{split}
\end{equation}
Consequently, the optimal bid from~(\ref{eq::denoisebid_bid_formula}) is directly proportional to this posterior expectation:
\begin{equation}
\label{eq::single_ctr_bid_formula}
        bid_t = \frac{CVR_t + qC}{p + q}\mathbb{E}\left[ CTR_t \middle| \widehat{CTR}_t \right].
\end{equation}
In turn, expectation is equal:
\begin{equation}
    \label{eq::ctr_only_expectatio}
    \begin{split}
        \mathbb{E}\big[ CTR_t \,\Big|\, \widehat{CTR}_t \big] = \frac{\int CTR_t \cdot p\big(\widehat{CTR}_t \big| CTR_t\big) p(CTR_t)\,\mathrm{d} CTR_t}{\int p\big(\widehat{CTR}_t \,\big|\, CTR_t\big) p(CTR_t) \,\mathrm{d} CTR_t}.
    \end{split}
\end{equation}

\paragraph{Noise model.} 
While we do not assume that estimation errors are strictly Gaussian in all practical scenarios, modeling noise as additive and normally distributed in the logit space is a well-motivated and tractable approximation~\cite{bishop2006pattern}
. 
This choice is particularly natural for CTR predictors, which typically generate a real-valued score before applying a sigmoid transformation. 
Furthermore, characterizing the logit distribution via its first two moments provides a standard basis for quantifying predictive uncertainty~\cite{kendall2017uncertainties}.

Formally, since click probabilities $CTR_t \in [0, 1]$, we perform the estimation in the logit space by defining the logit-CTR $\xi_t$ and its noisy observation $\hat{\xi}_t$:
\begin{equation}
    \label{eq::logit_ctrs}
    \xi_t = \ln \frac{CTR_t}{1 - CTR_t}, \quad \hat{\xi}_t = \ln \frac{\widehat{CTR}_t}{1 - \widehat{CTR}_t}.
\end{equation}
We assume the observation model as:
\begin{equation}
    \hat{\xi}_t = \xi_t + \varepsilon_t, \quad \varepsilon_t \sim \mathcal{N}\left(0, \sigma_t^2\right),
\end{equation}
where $\left\{\varepsilon_t\right\}_{t=1}^T$ are independent random variables and independent on the ground-truth logit-CTRs $\xi_t$.
Then the likelihood can be written as:
\begin{equation}
    \label{eq::logitgausmix_likelihood}
    p\left(\hat{\xi}_t \,\middle|\, \xi_t \right) = \frac{1}{\sqrt{2 \pi} \sigma_t} e^{-\frac{\left(\hat{\xi}_t - \xi_t\right)^2}{2 \sigma_t^2}}.
\end{equation}

\paragraph{Gaussian mixture prior.} 
Real-world CTR distributions are often complex and multimodal. 
Therefore, we model the prior distribution of the logit-CTR $\xi$ as a Gaussian Mixture:
\begin{equation}
    \label{eq::logit_gaus_mix_prior}
    \begin{split}
        &\xi \sim \sum_{k=1}^K \pi_k \cdot \mathcal{N}\left(\mu_k, \theta_k^2\right), \\
    \end{split}
\end{equation}
where $K$ is number of mixture components, $\pi_k$, $\mu_k$ and $\theta_k^2$ are the weight, mean and variance of the $k$-th component, respectively. 
Such a prior allows the framework to adapt to the empirical distribution of CTRs generated by the ML model trained on a given dataset.

To compute $\mathbb{E}\left[ CTR_t \middle| \widehat{CTR}_t \right]$~(\ref{eq::ctr_only_expectatio}), we first perform a change of variables to the logit space:
\begin{equation}
    \label{eq::ctr_expectation_through_logits}
    \begin{split}
        \mathbb{E}\Big[ CTR_t \Big| &\widehat{CTR}_t \Big] = \frac{\int \sigma(\xi_t) p\left(\hat{\xi}_t \middle| \xi_t\right) p(\xi_t) \,\mathrm{d}\xi_t}{\int p\left(\hat{\xi}_t \,\middle|\, \xi_t\right) p(\xi_t) \,\mathrm{d}\xi_t}.
    \end{split}
\end{equation}
Next, we substitute the likelihood~\eqref{eq::logitgausmix_likelihood} and the prior~\eqref{eq::logit_gaus_mix_prior} into~\eqref{eq::ctr_expectation_through_logits}:
\begin{equation}
    \label{eq::ctr_expectation_logitgausmix_logitgausnoise}
    \begin{split}
        \mathbb{E}\big[ CTR_t \,\big|\, \widehat{CTR}_t \big] = \frac{\sum_{k=1}^K \pi_k \int \sigma(\xi_t) \mathcal{N}\big(\hat{\xi}_t \,\big|\, \xi_t, \sigma_t^2\big) \mathcal{N}\big(\xi_t \,\big|\, \mu_k, \theta_k^2\big) \,\mathrm{d}\xi_t}{\sum_{k=1}^K \pi_k \int \mathcal{N}\big(\hat{\xi}_t \,\big|\, \xi_t, \sigma_t^2\big) \mathcal{N}\big(\xi_t \,\big|\, \mu_k, \theta_k^2\big) \,\mathrm{d}\xi_t},
    \end{split}
\end{equation}
where $\mathcal{N}\left(\, \cdot \,\middle|\, \mu, \sigma^2\right)$ denotes the Normal Probability Density Function with mean $\mu$ and variance $\sigma^2$.

By leveraging the property that the product of two Gaussian kernels is a scaled Gaussian and applying the probit approximation to the resulting sigmoid-Gaussian integrals (see Appendix~\ref{appx::ctr-only_exp_derivation} for the detailed derivation), we obtain a computationally efficient closed-form expression:
\begin{equation}
\label{eq::ctronly_logitgausmix_final_short}
\mathbb{E}\left[CTR_t \middle| \widehat{CTR}_t\right] \approx \sum_{k=1}^K \pi_{t,k}^\prime \sigma\left( \frac{\mu^\prime_{t,k}}{\sqrt{1 + \frac{\pi}{8}{\sigma^\prime_{t, k}}^2}} \right),
\end{equation}
where $\mu_{t, k}^\prime$ and ${\sigma_{t, k}^\prime}^2$ are the posterior mean and variance of the $k$-th component and $\pi_{t, k}^\prime \propto \pi_k \cdot \mathcal{N}(\hat{\xi}_t \mid \mu_k, \sigma_t^2 + \theta_k^2)$ are the posterior weights.

\paragraph{Empirically reconstructed prior.} 
Implementing the \texttt{DenoiseBid} requires the parameters $\pi_k, \mu_k, \theta_k$ of the prior distribution $p(\xi)$ introduced in \eqref{eq::logit_gaus_mix_prior}.
In the real world, we do not have access to ground-truth logits~$\xi_t$, so we can only use noisy estimates~$\hat{\xi}_t$ to approximate them.

To recover the prior distribution from noisy samples, we formulate the task as a density estimation problem with measurement errors. 
Since we assume the prior $p(\xi_t)$ is a Gaussian Mixture, the noise model $p\left(\hat{\xi}_t \mid \xi_t \right)$ is Gaussian, and noise is uncorrelated with CTR, the marginal distribution of the observed logits is also a Gaussian Mixture:
\begin{equation}
    \label{eq::marginal_dist_logits}
    \begin{split}
        p(\hat{\xi}_t) = \int p\big(\hat{\xi}_t \,\big|\, \xi_t\big)\cdot p(\xi_t) \,\mathrm{d}\xi_t =\sum_{k=1}^K \pi_k \mathcal{N}\left( \hat{\xi}_t \,\middle|\, \mu_k, \theta_k^2 + \sigma_t^2 \right),
    \end{split}
\end{equation}
where $\left\{\hat{\xi}_t, \sigma_t^2\right\}$ is CTR model output at auction $t$.

To estimate the parameters $\bm{\varphi} = \{ \pi_k, \mu_k, \theta_k^2 \}_{k=1}^K$, we employ the Extreme Deconvolution (XDGMM) technique~\cite{bovy2011extreme} that maximizes the following likelihood:
\begin{equation}
    \label{eq::ctr_only_likelihood}
    \begin{aligned}
        \mathcal{L}\Big(\bm{\varphi} \,\Big|\, \big\{\hat{\xi}_t\big\}_{t=1}^T, \,\left\{ \sigma^2_t \right\}_{t=1}^T\Big) = \sum_{t=1}^T \ln p\big(\hat{\xi}_t \,\big|\, \bm{\varphi}, \sigma_t^2\big) = \sum_{t=1}^T \ln \left( \sum_{k=1}^K \pi_k \mathcal{N}\left( \hat{\xi}_t \,\middle|\, \mu_k, \theta_k^2 + \sigma_t^2 \right) \right).
    \end{aligned}
\end{equation}
Unlike the standard EM algorithm, which treats observations as exact, XDGMM considers individual noise variances $\sigma_t^2$ for each sample.
Therefore, XDGMM is well-suited for reconstructing the CTR distribution for autobidding.


\subsection{Bidding under joint CTR-CVR uncertainty}
In the previous section, we assumed independence between CTR and CVR to simplify the derivation.  
In the case of joint uncertainty, which is closer to real-world scenarios, we consider correlations between CTR and CVR. 
To capture this relationship, we generalize the proposed framework to the \emph{joint} logit-space $\bm{\eta}_t = \begin{pmatrix}
    \xi_t & \zeta_t
\end{pmatrix}^\top$, where $\xi_t$ and $\zeta_t$ denotes logits of CTR and CVR, respectively:
\begin{equation}
    \label{eq::logit_joint}
    \xi_t = \ln\frac{CTR_t}{1 - CTR_t}, \quad 
    \zeta_t = \ln\frac{CVR_t}{1 - CVR_t}.
\end{equation}
In this setting, while we still assume independence between CTR and CVR across distinct auctions, we treat them as potentially correlated within each auction.
Consequently, the expectation of their product in~\eqref{eq::denoisebid_bid_formula} can no longer be factorized and requires the evaluation of:
\begin{equation}
    \label{eq::joint_expectations_general_form}
    \begin{split}
        &\mathbb{E}\big[CTR_t \,\big|\, \hat{\bm{\eta}}_t\big] = \textstyle\int \sigma(\xi_t) \, p(\bm{\eta}_t \,|\, \hat{\bm{\eta}}_t) \mathrm{d}^2\bm{\eta}_t, \\
        \mathbb{E}\big[C&TR_t \, CVR_t \,\big|\, \hat{\bm{\eta}}_t] = \textstyle\int \sigma(\xi_t) \sigma(\zeta_t) \, p(\bm{\eta}_t | \hat{\bm{\eta}}_t) \mathrm{d}^2\bm{\eta}_t,
    \end{split}
\end{equation}
where $\hat{\bm{\eta}}_t = \begin{pmatrix}
    \hat{\xi}_t & \hat{\zeta}_t
\end{pmatrix}^\top$ is the vector of logits of noisy CTR and CVR observations.

Similarly to the one-dimensional case, we express expectations in terms of the likelihood and the prior:
\begin{equation}
    \label{eq::joint_expectation}
    \begin{split}
        \mathbb{E}\big[X &\,\big|\, \hat{\bm{\eta}}_t] =\frac{\int X \cdot p\big(\hat{\bm{\eta}}_t \,\big|\, \bm{\eta}_t\big) p(\bm{\eta}_t) \,\mathrm{d}^2 \bm{\eta}_t}{\int p\big(\hat{\bm{\eta}}_t \,\big|\, \bm{\eta}_t\big) p(\bm{\eta}_t) \,\mathrm{d}^2 \bm{\eta}_t},
    \end{split}
\end{equation}
where latent variable $X$ denotes $CTR_t$ or  $CTR_t \cdot CVR_t$.

\paragraph{Noise model.} 
Following the one-dimensional case, we assume that the observation errors for both CTR and CVR are additive and Gaussian in the logit space:
\begin{equation}
    \label{eq::joint_noise_model}
        \hat{\bm{\eta}}_t = \bm{\eta}_t + \bm{\varepsilon}_t, \quad \bm{\varepsilon}_t \sim \mathcal{N}(\bm{0}, \bm{\Sigma}_t),
\end{equation}
where $\{\bm{\varepsilon}_t\}_{t=1}^T$ are independent two-dimensional random vectors and they are independent from~$\{\bm{\eta}_t\}_{t=1}^T$.
Then likelihood can be written as:
\begin{equation}
    \label{eq::joint_likelihood}
    p\big(\hat{\bm{\eta}}_t \,\big|\, \bm{\eta}_t\big) = \frac{1}{2 \pi\sqrt{|\bm{\Sigma}_t|}}e^{- \frac{1}{2}\left(\hat{\bm{\eta}}_t - \bm{\eta}_t\right)^\top \bm{\Sigma}_t^{-1} \left(\hat{\bm{\eta}}_t - \bm{\eta}_t\right)}.
\end{equation}

\paragraph{Gaussian mixture prior.} 
To capture the correlations between CTR and CVR, we model the joint prior distribution of the logit-vector as a bivariate Gaussian Mixture:
\begin{equation}
    \label{eq::joint_logit_gaus_mix_prior}
    \begin{split}
        \bm{\eta}_t \sim &\sum_{k=1}^K \pi_k \mathcal{N}\left(\bm{\mu}_k, \bm{\Theta}_k\right),\\
        p(\bm{\eta}_t) = \sum_{k=1}^K &\frac{\pi_k}{2\pi\sqrt{|\bm{\Theta}_k|}}e^{-\frac{1}{2}(\bm{\eta}_t - \bm{\mu}_k)^\top \bm{\Theta}^{-1}_k (\bm{\eta}_t - \bm{\mu}_k)},
    \end{split}
\end{equation}
where $K$ is number of mixture components, $\pi_k$, $\bm{\mu}_k$, and $\bm{\Theta}_k$ are the weight, mean vector, and covariance matrix of the $k$-th two-dimensional Gaussian.

Substituting the likelihood~\eqref{eq::joint_likelihood} and prior~\eqref{eq::joint_logit_gaus_mix_prior} into~\eqref{eq::joint_expectation} yields the following expression for expectation:
\begin{equation}
    \label{eq::jpint_expectation_through_gaus_sum}
    \begin{split}
        \mathbb{E}\big[X \,\big|\, \hat{\bm{\eta}}_t\big] =\frac{\sum_{k=1}^K \pi_k \int X \mathcal{N}(\hat{\bm{\eta}}_t \mid \bm{\eta}_t, \bm{\Sigma}_t) \mathcal{N}(\bm{\eta}_t \mid \bm{\mu}_k, \bm{\Theta}_k) \,\mathrm{d}^2\bm{\eta}_t}{\sum_{k=1}^K \pi_k \int \mathcal{N}(\hat{\bm{\eta}}_t \mid \bm{\eta}_t, \bm{\Sigma}_t) \mathcal{N}(\bm{\eta}_t \mid \bm{\mu}_k, \bm{\Theta}_k) \,\mathrm{d}^2\bm{\eta}_t}.
    \end{split}
\end{equation}
Similar to the previous case, we can apply the Gaussian distribution property and arrive at the following formula:
\begin{equation}
    \label{eq::joint_expectation_short}
    \mathbb{E}\big[X \,\big|\, \hat{\bm{\eta}}_t\big] = \sum_{k=1}^K \pi^\prime_{t,k} \int X \cdot \mathcal{N}(\bm{\eta}_t \,|\, \bm{\mu}_{t,k}^\prime, \bm{\Sigma}^\prime_{t, k}) \,\mathrm{d}^2 \bm{\eta}_t,
\end{equation}
where $\pi_{t,k}^\prime$ are the posterior weights and $\mu_{t,k}^\prime$, $\Sigma_{t, k}^\prime$ are the updated posterior mean vectors and covariance matrices for each component (see detailed derivation in Appendix~\ref{appx::joint_exp_derivation})

To compute the CTR expectation, we exploit the property that the marginal posterior of $\xi_t$ remains normal, which allows the use of the probit approximation for each component:
\begin{equation}
    \label{eq::joint_ctr_probit_short}
    \mathbb{E}\big[CTR_t \,\big|\, \hat{\bm{\eta}}_t\big] \approx \sum_{k=1}^K \pi^\prime_{t,k} \sigma \left( \frac{(\bm{\mu}^\prime_{t,k})_0}{\sqrt{1 + \frac{\pi}{8}(\bm{\Sigma}^\prime_{t,k})_{0,0}}} \right),
\end{equation}
where $(\,\cdot\,)_0$ and $(\,\cdot\,)_{0,0}$ denote the first element of the mean vector and the top-left element of the covariance matrix.

However, the value ($X = \sigma(\xi_t)\sigma(\zeta_t)$) involves a product of two sigmoids over a correlated distribution. 
We employ Gauss-Hermite quadrature~\cite{abramowitz1948handbook, jackel2005note} to compute expectation. To account for correlations, we use the Cholesky decomposition $\bm{\Sigma}_{t,k}^\prime = \mathbf{LL}^\top$ to transform the standard nodes to match the mean and covariance of each component:
\begin{equation}
    \label{eq::gauss_hermite_approx_short}
    \begin{split}
        \mathbb{E}\big[CTR_t \,CVR_t \,\big|\, \hat{\bm{\eta}}_t\big] \approx \sum_{k=1}^K \pi^\prime_{t,k} \sum_{i,j=1}^M \lambda_i \lambda_j \sigma\left(\xi^{(i,j)}_{t,k}\right) \sigma\left(\zeta^{(i,j)}_{t,k}\right),
    \end{split}
\end{equation}
where nodes $\bm{\eta}^{(i,j)}_{t,k} = \begin{pmatrix}
    \xi^{(i,j)}_{t,k} & \zeta^{(i,j)}_{t,k}
\end{pmatrix}^\top$ are obtained via the Cholesky decomposition of the posterior covariance matrix $\bm{\Sigma}_{t,k}^\prime = \mathbf{L}_{t,k}\mathbf{L}_{t,k}^\top$ (see details in Appendix~\ref{appx::gaushermite_int_approx}).

Since the sigmoid function is smooth, a small number of nodes (e.g., $M=5$) provides sufficient precision for bidding purposes. Empirically, further increasing of $M$ yields no significant accuracy gains.
The posterior covariance matrix is only $2\times2$, so this Cholesky decomposition can be computed analytically. 
Thus, this approximation scheme is fast and satisfies the time requirements of a real-time bidding system.

\paragraph{Empirically reconstructed prior.} 
As before, the prior parameters $\bm{\psi} = \{\pi_k, \bm{\mu}_k, \bm{\Sigma}_k\}_{k=1}^K$ introduced in \eqref{eq::joint_logit_gaus_mix_prior} must be estimated from the noisy observations~$\hat{\bm{\eta}}_t$. To perform this estimation, we utilize the property that the marginal distribution of the observed logit-vectors is a bivariate Gaussian Mixture:
\begin{equation}
    \label{eq::joint_logits_marginal}
    \begin{split}
        \hat{\bm{\eta}}_t \sim \sum_{k=1}^K \pi_k \mathcal{N}(\bm{\mu}_k, \bm{\Theta}_k + \bm{\Sigma}_t).
    \end{split}
\end{equation}
where $\bm{\Sigma}_t$ is the noise covariance provided by the CTR and CVR models (typically diagonal, though \texttt{DenoiseBid} can also incorporate error correlations). 
We infer $\bm{\psi}$ by maximizing the log-likelihood via the XDGMM~\cite{bovy2011extreme}:
\begin{equation}
    \label{eq::joint_log_likelihood}
    \begin{split}
        \mathcal{L}(\bm{\psi} \mid \{\hat{\bm{\eta}}_t\}_{t=1}^T, \{\bm{\Sigma_t}\}_{t=1}^T) = \sum_{t=1}^T \ln p\left(\hat{\bm{\eta}}_t \mid \bm{\psi}, \bm{\Sigma}_t\right) = \sum_{t=1}^T & \ln \left( \sum_{k=1}^K \pi_k \mathcal{N}\left(\hat{\bm{\eta}}_t \mid \bm{\mu}_k, \bm{\Theta}_k + \bm{\Sigma}_t \right)\right).
    \end{split}
\end{equation}
Note that XDGMM can process data in parallel or in the background regime to avoid redundant overhead.
This feature makes it tractable for deployment in the real-time bidding system.

\section{Computational experiment}
\label{sec:experiment}
In this section, we evaluate the \texttt{DenoiseBid} framework against two baselines: the non-robust strategy~\cite{yang2019bid} and the Robust Optimization approach~\cite{pudovikov2025robust}. 
We use \texttt{DenoiseBid} under two prior distributions in the logit space: a normal distribution and a two-component Gaussian Mixture.
For noise modeling, we use two settings: synthetic noise and empirical noise extracted from the pre-trained CTR model.  
To reproduce the presented results, you can find the source code in the repository: \url{https://anonymous.4open.science/r/denoise-bid-uai26-1A60/}.


\paragraph{Setup.} 
We use an offline setting for comparison of the considered algorithms.
To ensure stable behavior patterns, we use relative budget and CPC constraints defined by fixed scaling factors $k_B$ and $k_C$ across campaigns:
\begin{equation}
    \label{eq::rel_exp_constraints}
    B= k_B \sum_{t=1}^T wp_t, \quad C= k_C \frac{\sum_{t=1}^T wp_t}{\sum_{t=1}^T CTR_t}.
\end{equation}
Further, we use $k_B =0.2 $ and $ k_C = 0.2$ to define budget and CPC upper bounds.

\paragraph{Datasets.} 
First, we use fully-synthetic data to verify the mathematical consistency of our CTR-only and joint methods. 
Then, we move to real-world cases using iPinYou and BAT datasets with empirical winning prices and model predictions. 
Finally, we test \texttt{DenoiseBid} on the Criteo Attribution dataset.
We train a CatBoost CTR model on this dataset and use CatBoost Virtual Ensembles to estimate knowledge uncertainty in CTR values.
This setup does not include any sampling of synthetic noise or data.  

\begin{table}[!ht]
  \centering
  \begin{threeparttable}
  \caption{Summary of experimental settings across datasets. The red cross indicates that the dataset admits such a type of noise; however, we skip this setting. Dash indicates that the noise type is infeasible for the dataset.}
  \label{table:experiments}
  \begin{tabular}{lcc}
    \toprule
    \textbf{Dataset} & \textbf{Synthetic noise}  & \textbf{Empirical noise} \\
    \midrule
    Synthetic & 
      \cmark & 
      \textemdash \\
    \addlinespace
    iPinYou & 
      \cmark & 
      \xmark \\
    \addlinespace
     BAT & 
      \cmark & 
      \textemdash \\
    \addlinespace
    Criteo Attribution & 
      \xmark & 
      \cmark \\
    \bottomrule
  \end{tabular}
  \end{threeparttable}
\end{table}

\subsection{Experiments with synthetic noise}
These experiments verify the mathematical consistency of our methods using known noise structures.

\paragraph{Datasets and setup.} 
We evaluate our approach on three datasets. 
In the \textbf{Synthetic} dataset, winning prices are generated from a log-normal distribution, and CTR/CVR logits are generated from a Gaussian mixture. 
For \textbf{iPinYou}, we use empirical winning prices, and CTR/CVR logits are generated from a Gaussian mixture. 
Finally, the \textbf{BAT} dataset leverages both real-world winning prices and model predictions of CTR/CVR directly from historical auction logs. 
Details of the experiment are provided in the Appendix~\ref{appx::synth_noise_experiment_details}.

\paragraph{Evaluation metrics.} 
To assess performance and constraint adherence, we define two key indicators:
\begin{itemize}
    \item The ratio of the optimal value for objective functions in problem~(\ref{eq::stochastic_optimization_problem}) to the theoretical optimum from problem~(\ref{eq::nonrobust_problem}) denoted as $R/R^*$.
    \item $\overline{CPC} / \overline{CPC}_{\text{camp}}$: the actual average cost-per-click normalized by the campaign's mean CPC. Since the target limit is set as $C = k_C \cdot \overline{CPC}_{\text{camp}}$, a strategy is considered cost-compliant only if this ratio below the threshold $k_C$.
\end{itemize}

\begin{figure*}[!ht]
    \centering
    \includegraphics[width=1.\linewidth]{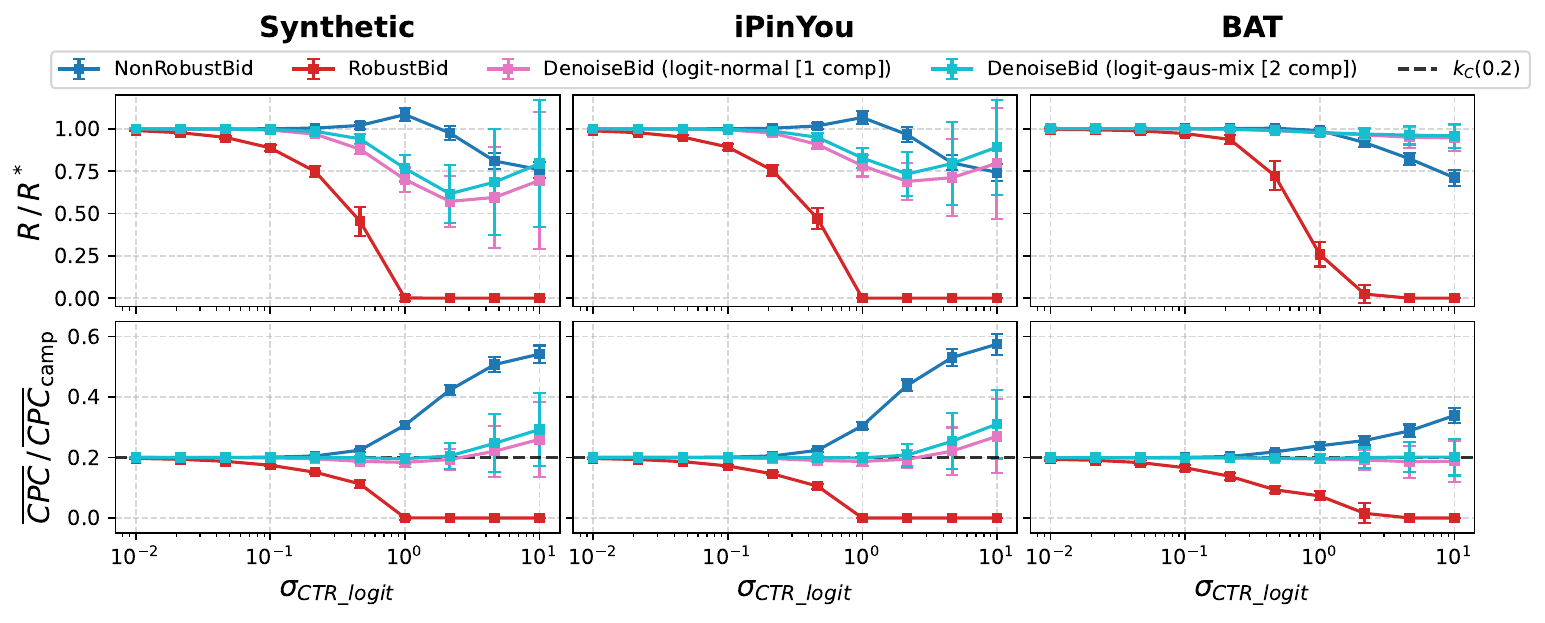}
    \caption{Performance comparison on the synthetic, iPinYou and BAT datasets under CTR uncertainty.}
    \label{fig::ctr-only_all_datasets}
\end{figure*}

\paragraph{Main results for CTR-only uncertainty.}
Figure~\ref{fig::ctr-only_all_datasets} illustrates the behavior of the strategies under increasing noise~$\sigma^\text{CTR}$. 
The non-robust baseline fails as noise increases, violating CPC limits and resulting in lower efficiency. 
While \texttt{RobustBid} remains CPC-compliant, it drops $R/R^*$. 
In contrast, \texttt{DenoiseBid} demonstrates the highest stability as noise increases, providing the best balance between conversion volume and constraint adherence.

\begin{figure*}[!ht]
    \centering
    \includegraphics[width=1.\linewidth]{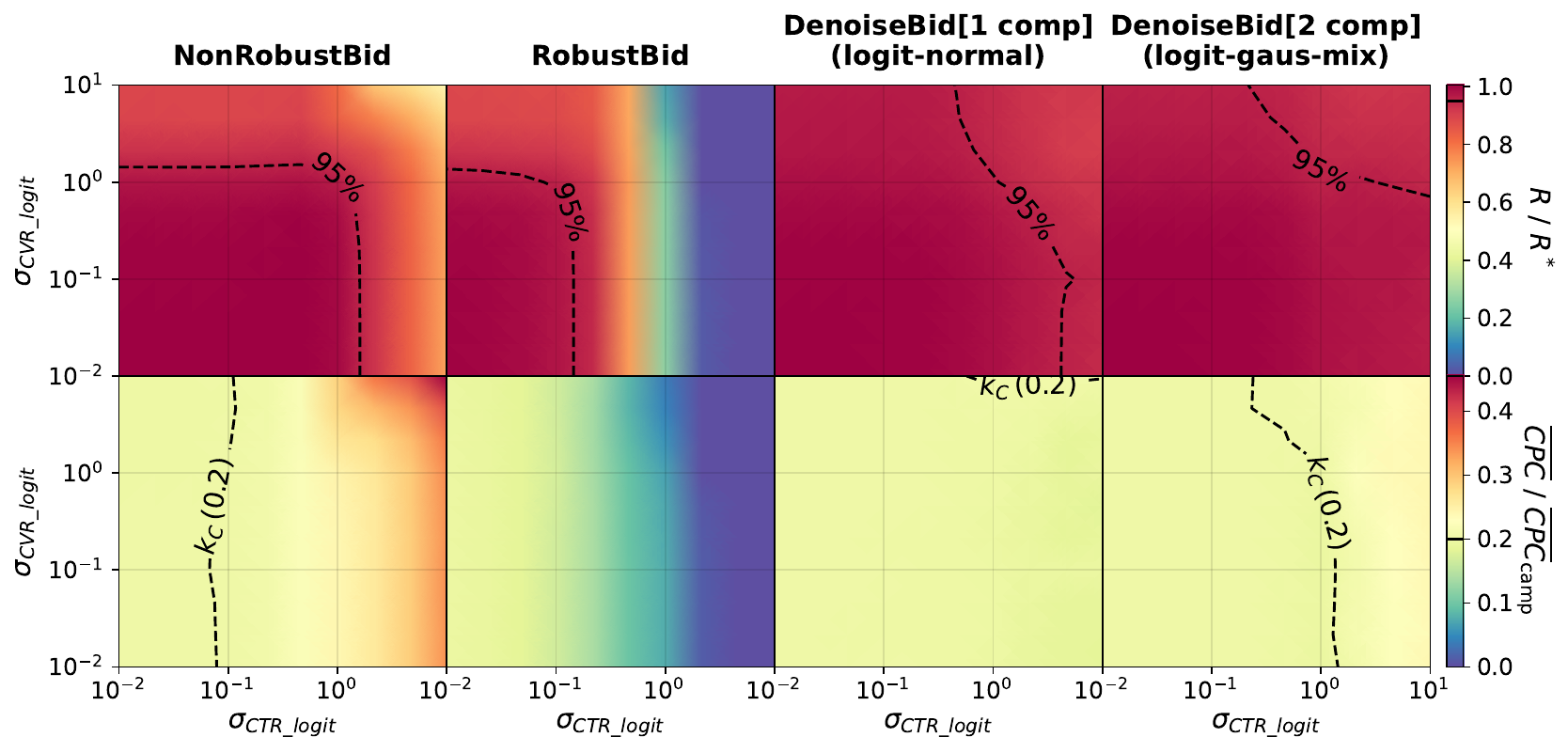}
    \caption{Performance comparison on the BAT dataset under joint CTR-CVR uncertainty.}
    \label{fig::joint_bat}
\end{figure*}

\paragraph{Main results for joint CTR-CVR uncertainty.}
Figure~\ref{fig::joint_bat} illustrates the metrics $R/R^*$ and $\overline{CPC} / \overline{CPC}_\text{camp}$ heatmaps under bivariate noise for the BAT dataset. 
The non-robust strategy fails to maintain cost compliance across the noise range, while \texttt{RobustBid} achieves compliance at the cost of significantly lower $R/R^*$. 
In contrast, \texttt{DenoiseBid} maintains a stable, near-optimal conversion volume while almost strictly adhering to constraints.
Similar heatmaps for the Synthetic and iPinYou datasets are shown in Appendix~\ref{sec::ipinsynth_heatmaps}.

\subsection{Experiments with empirical noise}
This setup evaluates actual realized outcomes from \textbf{Criteo Attribution} logs.

\paragraph{Methodology.} 
We trained CatBoost CTR and CVR models using Virtual Ensembles to obtain both predictions and logit variances. 
We considered two scenarios to simulate varying noise levels: 
\begin{enumerate}
    \item \textbf{Feature removal}, where features are excluded by importance increasing as a proxy for uncertainty,
    \item \textbf{Data scaling}, where training set size reduces (from 100\% to 0.1\%) to increase knowledge uncertainty.
\end{enumerate}
Details of this procedure are presented in Appendix~\ref{appx::empirical_noise_experiment_details}.

\paragraph{Metrics.} 
In contrast to the synthetic case, we evaluate uplifts using the following metrics:
\begin{itemize}
    \item relative error for number of conversions: $(conv - conv_{nr}) / conv_{nr}$, where $nr$ denotes the non-robust baseline.
    \item $(\overline{CPC} - \overline{CPC}_{nr}) / C$, representing the cost change normalized by the target $C$.
\end{itemize}

\begin{figure*}[!ht]
    \centering
    \includegraphics[width=1.\linewidth]{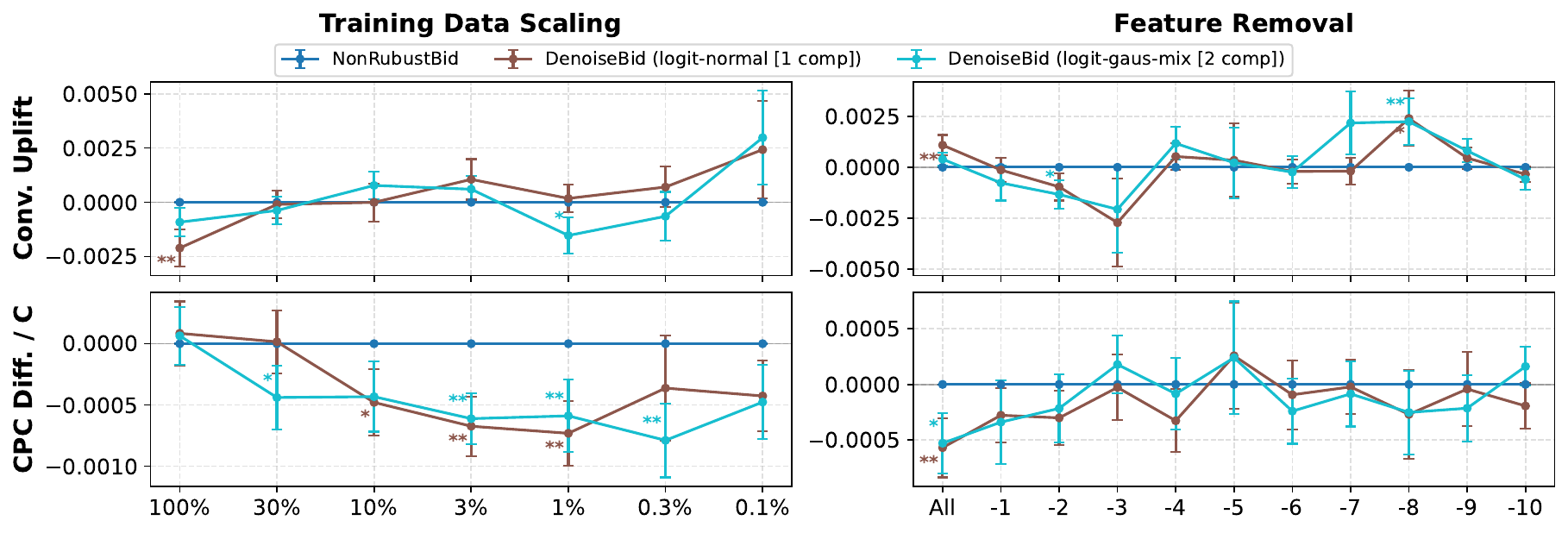}
    \caption{Relative uplifts in conversions and CPC deviation on the Criteo Attribution dataset. Markers * and ** denote statistical significance at the 10\% and 5\% levels, respectively.}
    \label{fig::criteo}
\end{figure*}

\paragraph{Main results.} Figure~\ref{fig::criteo} summarizes the relative uplifts. 
The logit-normal configuration provides a statistically significant ($p < 0.05$) reduction in CPC deviation in the 1\% and 3\% data scaling regimes. 
The variant with Gaussian mixture in logit space demonstrates broader cost stability, yielding significant CPC improvements across the 0.3\%, 1\%, and 3\% data regimes. 
Furthermore, it achieves significant conversion gains in the feature removal scenario when the 8 least informative variables are excluded.
The plots include error bars representing the Standard Error of the Mean to demonstrate the statistical significance of the observed performance gains.

\section{Limitations and future work}
We provide experiments only with GBDT models for conversion prediction; however, they are likely to become less popular as deep learning approaches become increasingly widespread. 
Therefore, we plan to adapt state-of-the-art models from~\cite{zhu2022bars} for our \texttt{DenoiseBid} framework. 
The overall framework will remain the same, with the main difference being in the specific steps taken to handle uncertainty.


\section{Conclusion}

We have proposed \texttt{DenoiseBid}, a Bayesian autobidding method that explicitly accounts for uncertainty in CTR and CVR estimates. 
The core idea is to replace the deterministic LP-based bidding formulation with a stochastic optimization problem in which bids are computed from posterior expectations of click and conversion probabilities.
Our approach makes three main contributions.
First, we formulated the autobidding problem under noisy CTR and CVR values and derived a closed-form bidding rule based on Bayesian posterior expectations.
Second, we developed a practical pipeline that recovers the prior distribution of CTR and CVR from noisy observations via Extreme Deconvolution (XDGMM) and computes denoised bids in closed form.
Third, we conducted extensive experiments on four datasets: Synthetic, iPinYou, BAT, and Criteo Attribution under both synthetic and empirical noise settings.
The results consistently show that \texttt{DenoiseBid} maintains near-optimal conversion volume while satisfying CPC constraints, outperforming both the non-robust baseline and the \texttt{RobustBid} strategy across a wide range of noise levels.

\appendix
\section*{Appendix}
\section{Solution of a stochastic optimization problem}
\label{appx::stoc_opt_prob}
In this section, we provide the detailed derivation of the optimal bidding policy for the stochastic optimization problem. We start with the linearized primal formulation of the \texttt{DenoiseBid} problem:

\begin{equation}
\label{eq::appx_primal}
    \begin{aligned}
        \max_{x_{t}} \quad & \sum_{t=1}^T x_t v_t, \\
        \text {s.t. } &\quad \sum _{t=1}^T x_{t} \cdot w p_{t} \leq B, \\
        & \sum_{t=1}^T x_{t} \cdot \left(wp_{t} - C \cdot c_t \right) \leq 0, \\
        \text{where}\quad& 0 \leq x_t \leq 1, \quad \forall t,
    \end{aligned}
\end{equation}
where $v_t = \mathbb{E}[CTR_t \cdot CVR_t \mid \mathcal{O}]$ and $c_t = \mathbb{E}[CTR_t \mid \mathcal{O}]$ are the posterior means of the conversion value and click probability, respectively.

\paragraph{Lagrangian Formulation.} To solve problem \eqref{eq::appx_primal}, we derive its Lagrangian dual. We introduce non-negative dual variables $p \geq 0$ and $q \geq 0$ associated with the budget and CPC constraints, respectively. 
To handle the box constraints $x_t \leq 1$, we also introduce auxiliary dual variables $r_t \geq 0$ for each auction. 
The Lagrangian $\mathcal{L}$ is formulated as follows:
\begin{equation}
\label{eq::denoisebid_lagrangian}
\begin{aligned}
    &\mathcal{L}(\mathbf{x}, p, q, \mathbf{r}) = \sum_{t=1}^T x_t v_t - p \left( \sum_{t=1}^T x_t wp_t - B \right) - \\
    &- q \left( \sum_{t=1}^T x_t (wp_t - C \cdot c_t) \right) - \sum_{t=1}^T r_t (x_t - 1),
\end{aligned}
\end{equation}
where $c_t = \mathbb{E}\left[CTR_t \middle| \mathcal{O}\right]$, $v_t = \mathbb{E}\left[CTR_t \cdot CVR_t \middle| \mathcal{O}\right]$ denote the posterior means of click probability and value at auction $t$, respectively. 
By rearranging terms to group the variables $x_t$, the dual objective is obtained by minimizing the Lagrangian with respect to $\mathbf{x}$.
\begin{equation}
    \label{eq::denoisebid_dual_problem}
    \begin{split}
        \min_{p, q, r_t} \quad & B \cdot p + \sum_{t = 1}^T r_t, \\
        \text{s.t.} \quad & wp_t \cdot p + (wp_t -  C \cdot c_t) \cdot q + r_t \geq v_t, \quad \forall t, \\
        \text{where} \quad & p \geq 0, q \geq 0, r_t \geq 0.
    \end{split}
\end{equation}

\paragraph{Optimality Conditions.} According to the Theorem of Complementary Slackness, the optimal primal solutions $\textbf{x}^*$ and dual solutions $p^*$, $q^*$ and $\mathbf{r}^*$ must satisfy:
\begin{equation}
\label{eq::complementary_1}
    x_t^* \cdot (v_t - wp_t \cdot p^* - (wp_t - C \cdot c_t) \cdot q^* - r_t^*) = 0,
\end{equation}
\begin{equation}
\label{eq::complementary_2}
    (x^*_t - 1) \cdot r^*_t = 0.
\end{equation}
We define the optimal bid price $bid_t^*$ as:
\begin{equation}
\label{eq::appx_denoisebid_bid_formula}
    bid^*_t = \frac{1}{p^* + q^*} v_t + \frac{q^*}{p^* + q^*} C \cdot c_t,
\end{equation}
By substituting \eqref{eq::appx_denoisebid_bid_formula} into the condition \eqref{eq::complementary_1}, we obtain:
\begin{equation}
\label{eq::transformed_slackness1}
    x^*_t \cdot \left( (bid^*_t - wp_t)(p^* + q^*) - r^*_t \right) = 0.
\end{equation}
Then, as it follows from \eqref{eq::complementary_2} and \eqref{eq::transformed_slackness1},
\begin{itemize}
    \item if the bidding agent wins the auction ($x_t^* > 0$), then $ (bid^*_t - wp_t)(p^* + q^*) - r^*_t = 0$. Thus, since $p^*, q^*, r_t^* \geq 0$, it must be $bid^*_t \geq wp_t$.
    \item if bidding agent loses the auction ($x_t^*$ = 0), then $r_t^* = 0$. Therefore, from dual condition $wp_t \cdot p + (wp_t - C \cdot c_t) \cdot q \geq v_t$, and it can be rewritten as $wp_t \geq bid_t^*$.
\end{itemize}

As a result, if $p$ and $q$ are the optimal solution of dual problem \eqref{eq::denoisebid_dual_problem}, optimal bidding formula is:

\begin{equation}
    bid_t = \frac{1}{p + q}\mathbb{E}\left[CTR_t \cdot CVR_t \middle| \mathcal{O} \right] + \frac{q}{p + q} C \cdot \left[CTR_t \middle| \mathcal{O} \right].
\end{equation}

\section{Derivation of the calculation formulas for expectations}

\subsection{CTR-only uncertainty}
\label{appx::ctr-only_exp_derivation}
\paragraph{Reduction of Gaussian Products.} To simplify the expression:
\begin{equation}
    \label{eq::ctr_expectation_logitgausmix_logitgausnoise_in_appdx}
    \begin{split}
        &\mathbb{E}\left[ CTR_t \middle| \widehat{CTR}_t \right] = \frac{\sum_k \pi_k \int \sigma(\xi_t) \cdot \mathcal{N}\left(\hat{\xi}_t \,\middle|\, \xi_t, \sigma_t^2\right) \mathcal{N}\left(\xi_t \,\middle|\, \mu_k, \theta_k^2\right)  \,\mathrm{d}\xi_t}{\sum_k \pi_k \int \mathcal{N}\left(\hat{\xi}_t \,\middle|\, \xi_t, \sigma_t^2\right) \mathcal{N}\left(\xi_t \,\middle|\, \mu_k, \theta_k^2\right) \,\mathrm{d}\xi_t},
    \end{split}
\end{equation}
we leverage a fundamental property of Normal densities: the product of two Gaussian kernels is itself a scaled Gaussian kernel:
\begin{equation}
    \label{eq::gaussian_product}
    \begin{split}
        \mathcal{N}\left(\hat{\xi}_t \,\middle|\, \xi_t, \sigma_t^2\right) \cdot \mathcal{N}\big(\xi_t \big| \mu_k, \theta_k^2\big) = \alpha_{t,k} \cdot \mathcal{N}\left(\xi_t \,\middle|\, \mu^\prime_{t,k}, {\sigma^\prime_{t,k}}^2\right),
    \end{split}
\end{equation}
where scaling factor $\alpha_{t, k} = \mathcal{N}\left(\hat{\xi}_t \,\middle|\, \mu_k, \sigma_t^2 + \theta_k^2 \right)$ represents the evidence of the $k$-th component, and
\begin{equation}
\label{eq::updated_params}
\begin{split}
    \alpha&_{t, k} = \frac{1}{\sqrt{2 \pi \left(\sigma_t^2 + \theta_k^2\right)}} e^{-\frac{\left(\hat{\xi}_t - \mu_k\right)^2}{2\left(\sigma^2_t + \theta_k^2 \right) } },\\
    &{\sigma^\prime_{t, k}}^2 = \frac{1}{1/\sigma_t^2 + 1/\theta_k^2}, \\
    \mu&^\prime_{t, k} = {\sigma^\prime_{t, k}}^2 \left( \frac{\hat{\xi}_t}{\sigma_t^2} + \frac{\mu_k}{\theta_k^2} \right).
\end{split}
\end{equation}
Substituting the Gaussian product property \eqref{eq::gaussian_product} into the expectation \eqref{eq::ctr_expectation_logitgausmix_logitgausnoise_in_appdx}, we observe that the posterior expectation is a weighted average of sigmoid-Gaussian integrals:
\begin{equation}
    \label{eq::posterior_sum_integrals}
    \begin{split}
        \mathbb{E}\Big[ &CTR_t \Big| \widehat{CTR}_t \Big] = \frac{\sum_k \pi_k \alpha_{t,k} \int \sigma(\xi_t) \mathcal{N}\left(\xi_t \,\middle|\, \mu^\prime_{t,k}, {\sigma^\prime_{t,k}}^2\right) \,\mathrm{d}\xi_t}{\sum_k \pi_k \alpha_{t,k}}.
    \end{split}
\end{equation}
\paragraph{Probit approximation.} The integral of a sigmoid with a Gaussian density does not have an exact closed-form solution in terms of elementary functions. To obtain a computationally efficient result, we utilize the probit approximation:
\begin{equation}
\int \sigma(\xi) \mathcal{N}\left(\xi \,\middle|\, \mu, \sigma^2\right) \,\mathrm{d}\xi \approx \sigma \left( \frac{\mu}{\sqrt{1 + \frac{\pi}{8}\sigma^2}} \right).
\end{equation}
Applying this approximation to each component in \eqref{eq::posterior_sum_integrals}, we present the final closed-form expression for the denoised CTR:
\begin{equation}
    \label{eq::ctronly_logitgausmix_final}
    \mathbb{E}\left[CTR_t \middle| \widehat{CTR}_t\right] \approx \sum_k \pi_{t,k}^\prime \sigma\left( \frac{\mu^\prime_{t,k}}{\sqrt{1 + \frac{\pi}{8}{\sigma^\prime_{t, k}}^2}} \right),
\end{equation}
where $\mu^\prime_{t,k}$ and ${\sigma^\prime_{t, k}}^2$ can be computed via \eqref{eq::updated_params}, and $\pi_{t,k}^\prime$ denotes posterior probability of $k$-th Gaussian given by $t$-th observation:
\begin{equation}
    \pi_{t,k}^\prime = \frac{\pi_k \alpha_{t,k}}{\sum_k \pi_k \alpha_{t,k}}.
\end{equation}

\subsection{Joint uncertainty}
\label{appx::joint_exp_derivation}

\paragraph{Reduction of Bivariate Gaussian Products.} We have an expression:
\begin{equation}
    \label{eq::appx_joint_expectation_through_gaus_sum}
    \mathbb{E}\big[X \,\big|\, \hat{\bm{\eta}}_t\big] = \frac{\sum_k \pi_k \int X \cdot \mathcal{N}(\hat{\bm{\eta}}_t \mid \bm{\eta}_t, \bm{\Sigma}_t) \mathcal{N}(\bm{\eta}_t \mid \bm{\mu}_k, \bm{\Theta}_k) \,\mathrm{d}^2\bm{\eta}_t}{\sum_k \pi_k \int \mathcal{N}(\hat{\bm{\eta}}_t \mid \bm{\eta}_t, \bm{\Sigma}_t) \mathcal{N}(\bm{\eta}_t \mid \bm{\mu}_k, \bm{\Theta}_k) \,\mathrm{d}^2\bm{\eta}_t}.
\end{equation}
By complete analogy with one-dimensional Gaussians \eqref{eq::gaussian_product}, the product of Normal densities is equal to Normal density with scaling factor:
\begin{equation}
\label{eq::2d_gaus_product}
    \mathcal{N}\left(\hat{\bm{\eta}}_t \,\big|\, \bm{\eta}_t, \bm{\Sigma}_t\right) \cdot \,\mathcal{N}\left(\bm{\eta}_t \mid \bm{\mu}_k, \bm{\Theta}_k\right) = \alpha_{t,k} \cdot \mathcal{N}\left(\bm{\eta}_t \,\big|\, \bm{\mu}^\prime_{t,k}, \bm{\Sigma}^\prime_{t,k}\right),
\end{equation}
where $\alpha_{t,k} = \mathcal{N}\left(\hat{\bm{\eta}}_t \,\big|\, \bm{\mu}_k, \bm{\Sigma}_t + \bm{\Theta}_k\right)$ is the evidence of the $k$-th Gaussian, and
\begin{equation}
    \label{eq::joint_updated_params}
    \begin{split}
        \alpha&_{t,k} = \frac{e^{-\frac{1}{2}\left(\hat{\bm{\eta}}_t - \bm{\mu}_k\right)^\top\left(\bm{\Sigma}_t + \bm{\Theta}_k\right)^{-1}\left(\hat{\bm{\eta}}_t - \bm{\mu}_k\right)}}{2\pi\sqrt{|\bm{\Sigma}_t + \bm{\Theta}_k|}}, \\
        &\bm{\Sigma}_{t,k}^\prime = \left(\bm{\Sigma}_t^{-1} + \bm{\Theta}_k^{-1} \right)^{-1}, \\
        \bm{\mu}&^\prime_{t,k} = \bm{\Sigma}_{t,k}^\prime\left(\bm{\Sigma}^{-1}_t \hat{\bm{\eta}}_t + \bm{\Theta}_k^{-1} \bm{\mu}_k\right).
    \end{split}
\end{equation}
Next, we substitute simplified form of Normal densities product \eqref{eq::2d_gaus_product} in posterior expectation expression \eqref{eq::appx_joint_expectation_through_gaus_sum}:
\begin{equation}
\label{eq::joint_expectation_reduced_form}
        \mathbb{E}\big[X \,\big|\, \hat{\bm{\eta}}_t\big] = \frac{\sum_k \pi_k \alpha_{t,k} \int X \cdot \mathcal{N}\left(\bm{\eta}_t \,\big|\, \bm{\mu}_{t, k}^\prime, \bm{\Sigma}^\prime_{t, k}\right)\,\mathrm{d}^2 \bm{\eta}_t}{\sum_k \pi_k \alpha_{t,k}},
\end{equation}
or, in closer form, with redefining posterior probability $\pi^\prime_{t,k} = \frac{\pi_k \alpha_{t, k}}{\sum_k \pi_k \alpha_{t, k}}$ of $k$-th Gaussian given by $t$-th estimates:
\begin{equation}
\label{eq::appx_joint_expectation_reduced_form2}
    \mathbb{E}\big[X \,\big|\, \hat{\bm{\eta}}_t\big] = \sum_k \pi^\prime_{t,k}\int X \cdot \mathcal{N}\left(\bm{\eta}_t \,\big|\, \bm{\mu}_{t,k}^\prime, \bm{\Sigma}^\prime_{t, k}\right)\,\mathrm{d}^2 \bm{\eta}_t.
\end{equation}

\subsection{Numerical approximation of integral}
\label{appx::gaushermite_int_approx}
\paragraph{Probit approximation for CTR expectation.} Now, we focus on approximating the integrals in \eqref{eq::appx_joint_expectation_reduced_form2}. For case with the click-through rate expectation $X = CTR_t = \sigma(\xi_t)$, the integral involves only a single sigmoid function over a bivariate Normal distribution. Since the marginal distribution of $\xi_t$ is still Gaussian, we can directly apply the probit approximation as in the one-dimensional case:
\begin{equation}
    \label{eq::joint_ctr_probit}
    \mathbb{E}\big[CTR_t \,\big|\, \hat{\bm{\eta}}_t\big] \approx\sum_k \pi^\prime_{t,k} \sigma \left( \frac{\left(\bm{\mu}^\prime_{t,k}\right)_0}{\sqrt{1 + \frac{\pi}{8}\left(\bm{\Sigma}^\prime_{t,k}\right)_{0,0}}} \right),
\end{equation}
where $(\,\cdot\,)_0$ and $(\,\cdot\,)_{0,0}$ denote the first element of vector and the top left element of matrix, respectively.

\paragraph{Numerical approximation via Gauss-Hermite quadrature for value expectation.} The expectation of the conversion value, when $X = CTR_t \cdot CVR_t = \sigma(\xi_t) \cdot \sigma(\zeta_t)$, involves the product of two sigmoid functions. Probit approximation has no multi-dimensional generalization, so, for calculate it with high computational efficiency while preserving accuracy, we employ Gauss-Hermite quadrature.

This method allows approximate convolution of function $f(\bm{\eta})$ with normal kernel by evaluating it at set $M \times M$ pre-defined nodes:
\begin{equation}
    \label{eq::gauss_hermite_approx}
    \int f(\bm{\eta}) \cdot \mathcal{N}\left(\bm{\eta} \mid \bm{\mu}, \bm{\Sigma}\right) \,\mathrm{d}^2 \bm{\eta} \approx \sum_{i=1}^M \sum_{j=1}^M \lambda_i \lambda_j f\left(\bm{\eta}^{(i,j)}\right),
\end{equation}
where the quadrature nodes $\bm{\eta}^{(i,j)}$ are obtained via the Cholesky decomposition of the posterior covariance matrix $\bm{\Sigma} = \mathbf{LL}^\top$ as:
\begin{equation}
    \bm{\eta}^{(i,j)} = \bm{\mu} + \sqrt{2} \mathbf{L} \begin{pmatrix} a_i \\ a_j \end{pmatrix},
\end{equation}
and $\{a_i\}_i$, $\{\lambda_i\}_i$ are the standard Gauss-Hermite nodes and weights.

Since the sigmoid function is smooth, a small number of nodes (e.g., $M=5$) provides sufficient precision for bidding purposes. The posterior covariance matrix is only $2\times2$, so, this Cholesky decomposition can be computed analytically. Thus, this approximation scheme is highly performant, satisfying the time requirements of RTB.

Therefore, for each auction $t$ separately, we can compute:
\begin{equation}
    \int \sigma(\xi_t) \sigma(\zeta_t) \cdot \mathcal{N}(\bm{\eta}_t \mid \bm{\mu}_{t, k}^\prime, {\bm{\Sigma}_{t, k}^\prime}) \,\mathrm{d}^2\bm{\eta}_t \approx \sum_{i=1}^M \sum_{j=1}^M \lambda_i \lambda_j  \cdot \sigma\left(\xi^{(i,j)}_{t,k} \right) \sigma\left(\zeta^{(i,j)}_{t,k} \right),
\end{equation}

\paragraph{Calculation formula.} As a result, we can approximate every integral in sum \eqref{eq::appx_joint_expectation_reduced_form2} and provide final expression:
\begin{equation}
    \mathbb{E}[CTR_t \cdot CVR_t \mid \bm{\eta}_t] \approx \sum_k \pi_{t,k}^\prime \sum_{i=1}^M \sum_{j=1}^M \lambda_i \lambda_j  \cdot \sigma\left(\xi^{(i,j)}_{t,k} \right) \sigma\left(\zeta^{(i,j)}_{t,k} \right),
\end{equation}
where $\pi^\prime_{t, k} = \frac{\pi_k \alpha_{t,k}}{\sum_k \pi_k \alpha_{t,k}}$,
\begin{equation}
    \begin{split}
        &\begin{pmatrix}
            \xi^{(i, j)}_{t,k} \\ \zeta^{(i, j)}_{t,k}
        \end{pmatrix} = \bm{\mu}_{t, k}^\prime + \sqrt{2} \mathbf{L}_{t,k} \begin{pmatrix}
            a_i \\ a_j
        \end{pmatrix}, \\
    & \quad \mathbf{L}_{t,k} = \begin{pmatrix} \sqrt{s_1} & 0 \\ s_0/\sqrt{s_1} & \sqrt{s_2 - s_0^2/s_1} \end{pmatrix}, \\
    \text{where } & \begin{pmatrix}
        s_1 & s_0 \\
        s_0 & s_2
    \end{pmatrix} = \bm{\Sigma}^\prime_{t, k}.
    \end{split}
\end{equation}
Parameters $\alpha_{t,k}, \bm{\mu}^\prime_{t,k}, \bm{\Sigma}^\prime_{t, k}$ defined in \eqref{eq::updated_params}, and $a_k$, $\lambda_k$ are tabular Gauss-Hermite nodes and weights.

\section{Experiment details}

In this section, we provide exhaustive details regarding the datasets, model architectures, and numerical settings used in our evaluation.

As was mentioned in Section \ref{sec:experiment}, we utilize the relative constraints:
\begin{equation}
    \label{eq::appx_rel_constraints}
    \begin{split}
        &B = k_B \cdot \sum_t wp_t, \\
        C &= k_C \cdot \frac{\sum_t wp_t}{\sum_t CTR_t}.
    \end{split}
\end{equation}
In all experiments $k_C=0.2$ and $k_B=0.2$ were used.

\subsection{Experiments with synthetic noise (synthetic, iPinYou, BAT)}
\label{appx::synth_noise_experiment_details}

\paragraph{Synthetic}
\begin{itemize}
    \item Number of auctions $T=1000$, number of campaigns $N=200$.
    \item $wp$ sampled from a log-normal distribution $\ln wp \sim \mathcal{N}(\mu_{wp}, \sigma_{wp}^2)$ with $\mu_{wp} = \ln 100.0 - \sigma_{wp}$, $\sigma_{wp} = 1.0$.
    \item CTR sampled from a log-normal distribution $\text{logit}\, CTR \sim \sum^K_{k=1} \pi_k \mathcal{N}(\mu_k, \sigma_k^2)$ with $K=3$, $\{\pi_1 = 0.6, \pi_2 = 0.2, \pi_3 = 0.2\}$, $\{\mu_1 = -3.0 ,\mu_2 = -2.5,\mu_3 = -1.0 \}$, $\{\sigma_1 = 0.3 ,\sigma_2 = 0.3,\sigma_3 = 0.4 \}$.
    \item CVR sampled from a log-normal distribution $\text{logit}\, CVR \sim \sum^K_{k=1} \pi_k \mathcal{N}(\mu_k, \sigma_k^2)$ with $K=2$, $\{\pi_1 = 0.6, \pi_2 = 0.4\}$, $\{\mu_1 = -2.0 ,\mu_2 = -1.0\}$, $\{\sigma_1 = 0.7 ,\sigma_2 = 0.4\}$.
    \item In experiment with CTR-only uncertainty in CTR logits sampled normal noise with standard deviation from $10^{-2}$ to $10^1$.
    \item In experiment with joint uncertainty also in CVR logits sampled normal noise with standard deviation from $10^{-2}$ to $10^1$, uncorrelated with noise in CTR logits.
    \item GMM was fitted via XDGMM on the subsample of 400 auctions in every campaign.The EM algorithm was run for a maximum of 300 iterations or until a log-likelihood tolerance of $2.0 \cdot 10^{-5}$ was reached.
    \item For the joint CTR-CVR uncertainty case, we used a $5\times5$ Gauss-Hermite quadrature grid.
\end{itemize}

\paragraph{iPinYou}
\begin{itemize}
    \item Number of auctions $T=1000$, number of campaigns $N=200$.
    \item $wp$ sampled from logs.
    \item CTR sampled from a log-normal distribution $\text{logit}\, CTR \sim \sum^K_{k=1} \pi_k \mathcal{N}(\mu_k, \sigma_k^2)$ with $K=3$, $\{\pi_1 = 0.6, \pi_2 = 0.2, \pi_3 = 0.2\}$, $\{\mu_1 = -3.0 ,\mu_2 = -2.5,\mu_3 = -1.0 \}$, $\{\sigma_1 = 0.3 ,\sigma_2 = 0.3,\sigma_3 = 0.4 \}$.
    \item CVR sampled from a log-normal distribution $\text{logit}\, CVR \sim \sum^K_{k=1} \pi_k \mathcal{N}(\mu_k, \sigma_k^2)$ with $K=2$, $\{\pi_1 = 0.6, \pi_2 = 0.4\}$, $\{\mu_1 = -2.0 ,\mu_2 = -1.0\}$, $\{\sigma_1 = 0.7 ,\sigma_2 = 0.4\}$.
    \item In experiment with CTR-only uncertainty in CTR logits sampled normal noise with standard deviation from $10^{-2}$ to $10^1$.
    \item In experiment with joint uncertainty also in CVR logits sampled normal noise with standard deviation from $10^{-2}$ to $10^1$, uncorrelated with noise in CTR logits.
    \item GMM was fitted via XDGMM on the subsample of 400 auctions in every campaign. The EM algorithm was run for a maximum of 300 iterations or until a log-likelihood tolerance of $2.0 \cdot 10^{-5}$ was reached.
    \item For the joint CTR-CVR uncertainty case, we used a $5\times5$ Gauss-Hermite quadrature grid.
\end{itemize}

\paragraph{BAT}
\begin{itemize}
    \item Number of campaigns $N=334$ (and in each campaign individual number of auctions).
    \item $wp$ sampled from logs.
    \item CTR sampled from logs.
    \item CVR sampled from logs.
    \item In experiment with CTR-only uncertainty in CTR logits sampled normal noise with standard deviation from $10^{-2}$ to $10^1$.
    \item In experiment with joint uncertainty also in CVR logits sampled normal noise with standard deviation from $10^{-2}$ to $10^1$, uncorrelated with noise in CTR logits.
    \item GMM was fitted via XDGMM on the subsample of 300 auctions in every campaign. The EM algorithm was run for a maximum of 300 iterations or until a log-likelihood tolerance of $2.0 \cdot 10^{-5}$ was reached.
    \item For the joint CTR-CVR uncertainty case, we used a $5\times5$ Gauss-Hermite quadrature grid.
\end{itemize}

\subsection{Experiments with empirical noise (Criteo Attribution)}
\label{appx::empirical_noise_experiment_details}
This experiment evaluates the framework's performance using actual model-derived uncertainty.

\paragraph{Scenarios:}
\begin{itemize}
    \item \textbf{Feature Removal:} We iteratively removed the top-N less important features, increasing predictive uncertainty.
    \item \textbf{Data Scaling:} The models were trained on subsets of the original data (100\%, 30\%, 10\%, 3\%, 1\%, 0.3\%, 0.1\%) while maintaining the same validation set for evaluation.
\end{itemize}

\paragraph{Model Training.} We trained \texttt{CatBoost} classifiers for both CTR and CVR tasks. We used the Virtual Ensembles mode ($\texttt{posterior\_sampling = True}$), which computes the variance of predictions across multiple sub-models in virtual ensemble.
\begin{itemize}
    \item \textbf{Features:} 12 features (11 of them categorical) including campaign ID, time since last click, and anonymized user metadata.
    \item \textbf{Uncertainty Quantification:} The output variance in logit space $\Sigma^2_t = \mathrm{diag}({\sigma^\text{CTR}_t}^2, {\sigma^\text{CVR}_t}^2)$ was used directly as the noise parameter for \texttt{DenoiseBid}.
\end{itemize}

\paragraph{Experiment specification:}
\begin{itemize}
    \item Number of campaigns $N=505$ (and in each campaign individual number of auctions).
    \item GMM was fitted via XDGMM on the subsample of 400 auctions in every campaign. The EM algorithm was run for a maximum of 300 iterations or until a log-likelihood tolerance of $2.0 \cdot 10^{-5}$ was reached.
\end{itemize}

\newpage
\section{Heat maps in experiments on synthetic and iPinYou datasets}
\label{sec::ipinsynth_heatmaps}
\begin{figure*}[!ht]
    \centering
    \includegraphics[width=1.\linewidth]{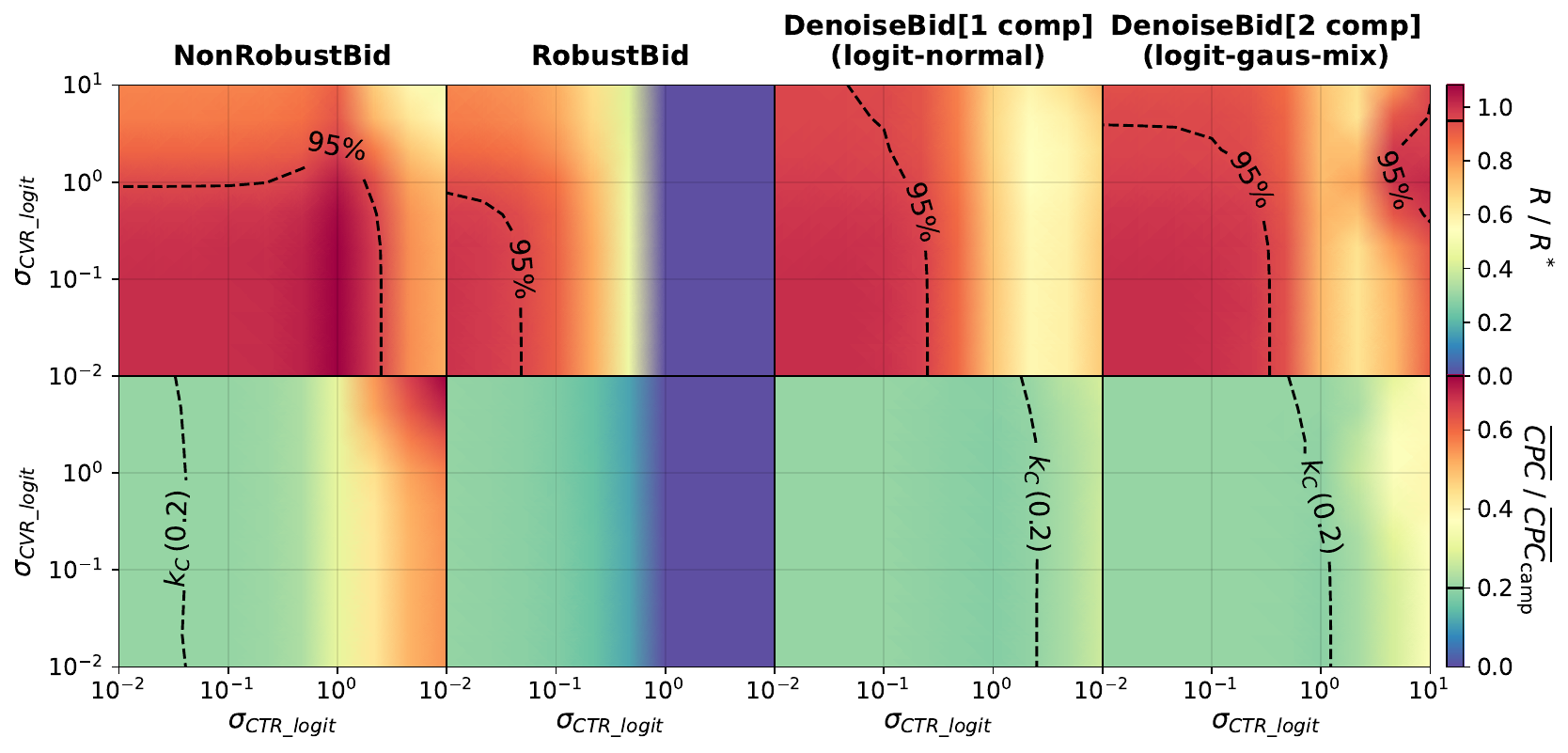}
    \caption{Performance comparison on the synthetic dataset under joint CTR-CVR uncertainty.}
    \label{fig::joint_synthetic}
\end{figure*}

\begin{figure*}[!ht]
    \centering
    \includegraphics[width=1.\linewidth]{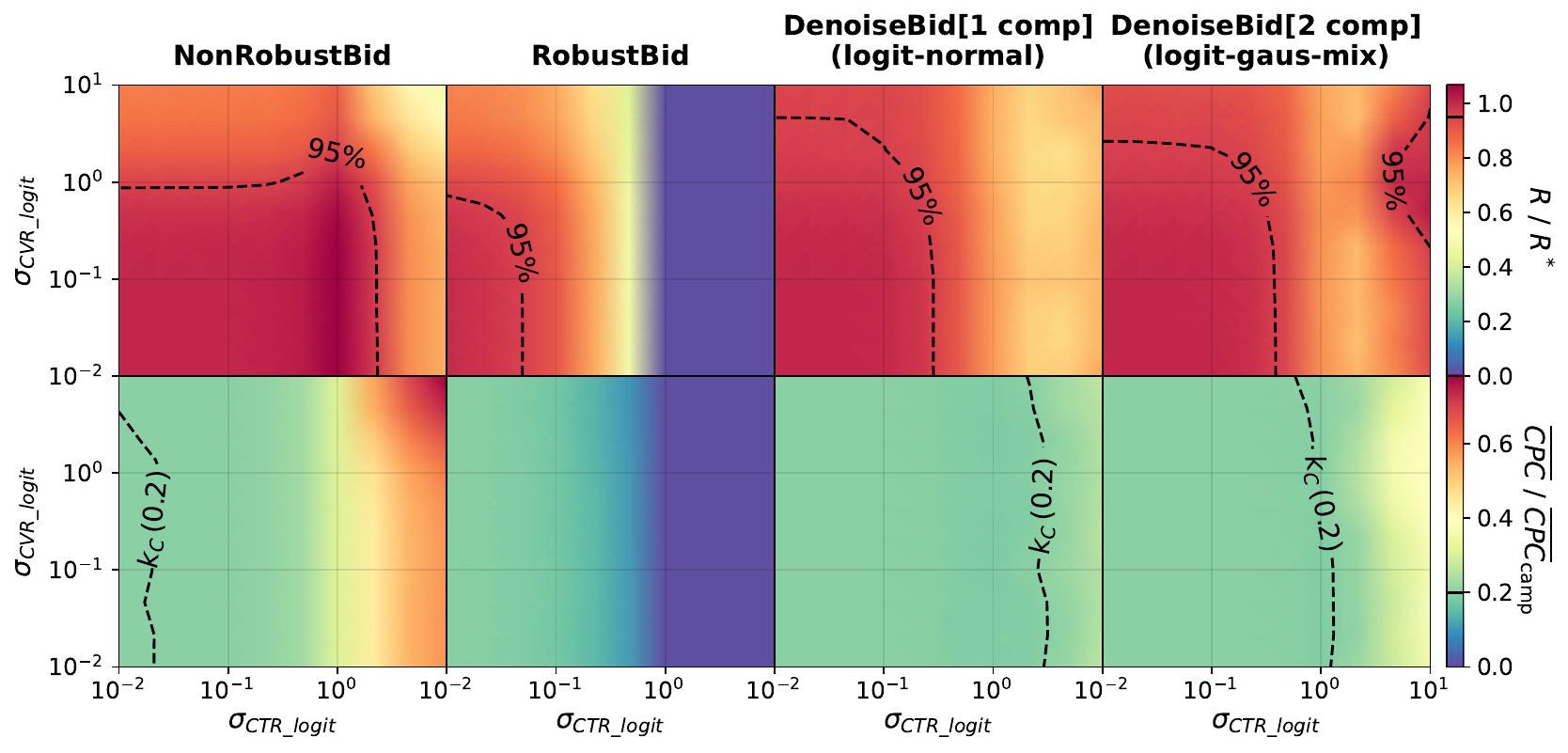}
    \caption{Performance comparison on the iPinYou dataset under joint CTR-CVR uncertainty.}
    \label{fig::joint_ipinyou}
\end{figure*}

This section provides visualizations for the joint CTR-CVR uncertainty scenarios on the Synthetic and iPinYou datasets. Figures~\ref{fig::joint_synthetic} and \ref{fig::joint_ipinyou} illustrate the metrics $R/R^*$ and $\overline{CPC} / \overline{CPC}_\text{camp}$ across a range of noise levels in the joint logit space ($\sigma^\text{CTR}$ and $\sigma^\text{CVR}$).

 

\begin{thebibliography}{19}
\providecommand{\natexlab}[1]{#1}
\providecommand{\url}[1]{\texttt{#1}}
\expandafter\ifx\csname urlstyle\endcsname\relax
  \providecommand{\doi}[1]{doi: #1}\else
  \providecommand{\doi}{doi: \begingroup \urlstyle{rm}\Url}\fi

\bibitem[Abdar et~al.(2021)Abdar, Pourpanah, Hussain, Rezazadegan, Liu, Ghavamzadeh, Fieguth, Cao, Khosravi, Acharya, et~al.]{abdar2021review}
Moloud Abdar, Farhad Pourpanah, Sadiq Hussain, Dana Rezazadegan, Li~Liu, Mohammad Ghavamzadeh, Paul Fieguth, Xiaochun Cao, Abbas Khosravi, U~Rajendra Acharya, et~al.
\newblock A review of uncertainty quantification in deep learning: Techniques, applications and challenges.
\newblock \emph{Information fusion}, 76:\penalty0 243--297, 2021.

\bibitem[Abramowitz and Stegun(1948)]{abramowitz1948handbook}
Milton Abramowitz and Irene~A Stegun.
\newblock \emph{Handbook of mathematical functions with formulas, graphs, and mathematical tables}, volume~55.
\newblock US Government printing office, 1948.

\bibitem[Aggarwal et~al.(2024)Aggarwal, Badanidiyuru, Balseiro, Bhawalkar, Deng, Feng, Goel, Liaw, Lu, Mahdian, et~al.]{aggarwal2024auto}
Gagan Aggarwal, Ashwinkumar Badanidiyuru, Santiago~R Balseiro, Kshipra Bhawalkar, Yuan Deng, Zhe Feng, Gagan Goel, Christopher Liaw, Haihao Lu, Mohammad Mahdian, et~al.
\newblock Auto-bidding and auctions in online advertising: A survey.
\newblock \emph{ACM SIGecom Exchanges}, 22\penalty0 (1):\penalty0 159--183, 2024.

\bibitem[Bishop and Nasrabadi(2006)]{bishop2006pattern}
Christopher~M Bishop and Nasser~M Nasrabadi.
\newblock \emph{Pattern recognition and machine learning}, volume~4.
\newblock Springer, 2006.

\bibitem[Bovy et~al.(2011)Bovy, Hogg, and Roweis]{bovy2011extreme}
Jo~Bovy, David~W Hogg, and Sam~T Roweis.
\newblock Extreme deconvolution: Inferring complete distribution functions from noisy, heterogeneous and incomplete observations.
\newblock \emph{The Annals of Applied Statistics}, 5\penalty0 (2B):\penalty0 1657, 2011.

\bibitem[Cai et~al.(2017)Cai, Ren, Zhang, Malialis, Wang, Yu, and Guo]{cai2017real}
Han Cai, Kan Ren, Weinan Zhang, Kleanthis Malialis, Jun Wang, Yong Yu, and Defeng Guo.
\newblock Real-time bidding by reinforcement learning in display advertising.
\newblock In \emph{Proceedings of the tenth ACM international conference on web search and data mining}, pages 661--670, 2017.

\bibitem[He et~al.(2021)He, Chen, Wu, Pan, Tan, Yu, Xu, and Zhu]{he2021unified}
Yue He, Xiujun Chen, Di~Wu, Junwei Pan, Qing Tan, Chuan Yu, Jian Xu, and Xiaoqiang Zhu.
\newblock A unified solution to constrained bidding in online display advertising.
\newblock In \emph{Proceedings of the 27th ACM SIGKDD Conference on Knowledge Discovery \& Data Mining}, pages 2993--3001, 2021.

\bibitem[J{\"a}ckel(2005)]{jackel2005note}
Peter J{\"a}ckel.
\newblock A note on multivariate gauss-hermite quadrature.
\newblock \emph{London: ABN-Amro. Re}, 2005.

\bibitem[Kendall and Gal(2017)]{kendall2017uncertainties}
Alex Kendall and Yarin Gal.
\newblock What uncertainties do we need in bayesian deep learning for computer vision?
\newblock \emph{Advances in neural information processing systems}, 30, 2017.

\bibitem[Khirianova et~al.(2025)Khirianova, Solodneva, Pudovikov, Osokin, Samosvat, Dorn, Ledovsky, and Zenkova]{khirianova2025bat}
Alexandra Khirianova, Ekaterina Solodneva, Andrey Pudovikov, Sergey Osokin, Egor Samosvat, Yuriy Dorn, Alexander Ledovsky, and Yana Zenkova.
\newblock Bat: Benchmark for auto-bidding task.
\newblock In \emph{Proceedings of the ACM on Web Conference 2025}, pages 2657--2667, 2025.

\bibitem[Malinin et~al.(2020)Malinin, Prokhorenkova, and Ustimenko]{malinin2020uncertainty}
Andrey Malinin, Liudmila Prokhorenkova, and Aleksei Ustimenko.
\newblock Uncertainty in gradient boosting via ensembles.
\newblock \emph{arXiv preprint arXiv:2006.10562}, 2020.

\bibitem[Ou et~al.(2023)Ou, Chen, Dai, Zhang, Liu, Tang, and Yu]{ou2023survey}
Weitong Ou, Bo~Chen, Xinyi Dai, Weinan Zhang, Weiwen Liu, Ruiming Tang, and Yong Yu.
\newblock A survey on bid optimization in real-time bidding display advertising.
\newblock \emph{ACM Transactions on Knowledge Discovery from Data}, 18\penalty0 (3):\penalty0 1--31, 2023.

\bibitem[Pudovikov et~al.(2025{\natexlab{a}})Pudovikov, Khirianova, Solodneva, Katrutsa, Samosvat, and Dorn]{pudovikov2025autobidding}
Andrey Pudovikov, Alexandra Khirianova, Ekaterina Solodneva, Aleksandr Katrutsa, Egor Samosvat, and Yuriy Dorn.
\newblock Autobidding arena: unified evaluation of the classical and rl-based autobidding algorithms.
\newblock \emph{arXiv preprint arXiv:2510.19357}, 2025{\natexlab{a}}.

\bibitem[Pudovikov et~al.(2025{\natexlab{b}})Pudovikov, Khirianova, Solodneva, Molodtsov, Katrutsa, Dorn, and Samosvat]{pudovikov2025robust}
Andrey Pudovikov, Alexandra Khirianova, Ekaterina Solodneva, Gleb Molodtsov, Aleksandr Katrutsa, Yuriy Dorn, and Egor Samosvat.
\newblock Robust autobidding for noisy conversion prediction models.
\newblock \emph{arXiv preprint arXiv:2510.08788}, 2025{\natexlab{b}}.

\bibitem[Shih et~al.(2023)Shih, Lai, and Huang]{shih2023robust}
Wen-Yueh Shih, Hsu-Chao Lai, and Jiun-Long Huang.
\newblock A robust real time bidding strategy against inaccurate ctr predictions by using cluster expected win rate.
\newblock \emph{IEEE Access}, 11:\penalty0 126917--126926, 2023.

\bibitem[Stram et~al.(2024)Stram, Abboud, Shtoff, Somekh, Raviv, and Koren]{stram2024mystique}
Rotem Stram, Rani Abboud, Alex Shtoff, Oren Somekh, Ariel Raviv, and Yair Koren.
\newblock Mystique: A budget pacing system for performance optimization in online advertising.
\newblock In \emph{Companion Proceedings of the ACM Web Conference 2024}, pages 433--442, 2024.

\bibitem[Yang et~al.(2019)Yang, Li, Wang, Wu, Tan, Xu, and Gai]{yang2019bid}
Xun Yang, Yasong Li, Hao Wang, Di~Wu, Qing Tan, Jian Xu, and Kun Gai.
\newblock Bid optimization by multivariable control in display advertising.
\newblock In \emph{Proceedings of the 25th ACM SIGKDD international conference on knowledge discovery \& data mining}, pages 1966--1974, 2019.

\bibitem[Zhang et~al.(2017)Zhang, Zhang, Rong, Ren, Li, and Wang]{zhang2017managing}
Haifeng Zhang, Weinan Zhang, Yifei Rong, Kan Ren, Wenxin Li, and Jun Wang.
\newblock Managing risk of bidding in display advertising.
\newblock In \emph{Proceedings of the Tenth ACM International Conference on Web Search and Data Mining}, pages 581--590, 2017.

\bibitem[Zhu et~al.(2022)Zhu, Dai, Su, Ma, Liu, Cai, Xiao, and Zhang]{zhu2022bars}
Jieming Zhu, Quanyu Dai, Liangcai Su, Rong Ma, Jinyang Liu, Guohao Cai, Xi~Xiao, and Rui Zhang.
\newblock Bars: Towards open benchmarking for recommender systems.
\newblock In \emph{Proceedings of the 45th International ACM SIGIR Conference on Research and Development in Information Retrieval}, pages 2912--2923, 2022.

\end{thebibliography}
\end{document}